\documentclass[conference]{IEEEtran}
\IEEEoverridecommandlockouts
\usepackage{cite}
\usepackage{amsmath,amssymb,amsfonts}
\usepackage{algorithmic}
\usepackage{graphicx}
\usepackage{textcomp}
\usepackage{xcolor}
\usepackage{romannum}
\usepackage{comment}
\usepackage{array}
\def\BibTeX{{\rm B\kern-.05em{\sc i\kern-.025em b}\kern-.08em
    T\kern-.1667em\lower.7ex\hbox{E}\kern-.125emX}}
\begin{document}

\title{PS8-Net: A Deep Convolutional Neural Network to Predict the Eight-State Protein Secondary Structure}
\author{
\IEEEauthorblockN{Md Aminur Rab Ratul, Maryam Tavakol Elahi, M. Hamed Mozaffari, WonSook Lee}
\IEEEauthorblockA{\textit{School of Electrical Engineering and Computer Science (SITE)}\\
\textit{University of Ottawa, Ottawa, Canada }\\
\{mratu076, mtava020, mmoza102, wslee\}@uottawa.ca}}
\maketitle

\begin{abstract}
Protein secondary structure is crucial to creating an information bridge between the primary and tertiary (3D) structures. Precise prediction of eight-state protein secondary structure (PSS) has significantly utilized in the structural and functional analysis of proteins in bioinformatics. Deep learning techniques have been recently applied in this research area and raised the eight-state (Q8) protein secondary structure prediction accuracy remarkably. Nevertheless, from a theoretical standpoint, there are still lots of rooms for improvement, specifically in the eight-state PSS prediction. In this study, we have presented a new deep convolutional neural network (DCNN), namely PS8-Net, to enhance the accuracy of eight-class PSS prediction. The input of this architecture is a carefully constructed feature matrix from the proteins sequence features and profile features. We introduce a new PS8 module in the network, which is applied with skip connection to extracting the long-term inter-dependencies from higher layers, obtaining local contexts in earlier layers, and achieving global information during secondary structure prediction. Our proposed PS8-Net achieves 76.89$\%$, 71.94$\%$, 76.86$\%$, and 75.26$\%$ Q8 accuracy respectively on benchmark CullPdb6133, CB513, CASP10, and CASP11 datasets. This architecture enables the efficient processing of local and global interdependencies between amino acids to make an accurate prediction of each class. To the best of our knowledge, PS8-Net experiment results demonstrate that it outperforms all the state-of-the-art methods on the aforementioned benchmark datasets.
\end{abstract}
\begin{IEEEkeywords}
Deep Convolutional Neural Network (DCNN), Protein Secondary Structure (PSS) Prediction, Bioinformatics, Local Context, Long-term
Interdependency, Skip Connection 
\end{IEEEkeywords}

\section{Introduction}
Protein is the vastly complex substance that is the basis of any living organism and involved in every part of cellular life. Proteins perform a colossal number of functions such as DNA replication, transporting molecules across membranes, giving structures to organisms, catalyzing chemical reactions, run the immune system, and regulating oxygenation. These are long-chain molecules made up of hundreds and thousands of 20 different kinds of smaller units called amino acids. It is necessary to know the protein's structure for understanding its functionalities. Reliably and accurately predicting protein's 3D structure from sequences of proteins is one of the most challenging problems in computational biology \cite{yaseen2014context} and is an essential step to identify functionality of those proteins at the molecular level. Prediction of protein's secondary structure is a vital step towards predicting protein tertiary (3D) structure \cite{zhou2014deep}, prediction of protein disorder \cite{heffernan2015improving}, and solvent accessibility prediction\cite{adamczak2004accurate}. 

In protein secondary structure (PSS), there are two regular PSS types: alpha-helix (H) and beta-strand (E), and one irregular PSS state: coil region (C). Predicting these three states, or the accuracy of these states is known as Q3 accuracy. According to hydrogen-bonding patterns, Sander has developed the Dictionary of Secondary Structure of Proteins (DSSP) to classify protein's secondary structure into eight states \cite{kabsch1983dictionary}. Furthermore, the DSSP algorithm designated helix into three types (G or 3-10 helix, H or $\alpha$-helix, and I or $\pi$-helix), two kinds for strand (E or $\beta$-strand, and B or $\beta$-bridge), and coil into three states (T or hydrogen-bonded turn, S or high curvature loop, and L or, irregular). The performance measure of these eight states is the Q8 accuracy (the eight-state protein secondary structure classification accuracy).

Previously many methods have extensively studied prediction of protein's secondary structures, including support vector machine techniques (SVM) \cite{hua2001novel}\cite{guo2004novel}, artificial neural networks\cite{qian1988predicting,jones1999protein,buchan2013scalable}, hidden Markov model\cite{asai1993prediction}\cite{aydin2006protein}, bidirectional recurrent neural network (BRNN)\cite{baldi1999exploiting,chen2007cascaded,mirabello2013porter,torrisi2018porter}, and the probability graph models\cite{schmidler2000bayesian,chu2004graphical}. Because of the advancement of deep learning, recently, most of the protein secondary structure prediction models based on deep learning, and these models yield a promising outcomes\cite{fang2018mufold,zhou2018cnnh_pss,guo2019deepaclstm,guo2018protein,wang2016protein,li2016protein}. Most of the methods of protein secondary structure prediction (PSS) have been heavily focused on three-state (Q3) classification. However, predicting the eight-class protein secondary structure is more challenging and complex, and reveals more detailed local information. Additionally, only a few deep learning architectures exist which have been build to predict eight-state PSS. Therefore, in this paper, we only emphasize on predicting eight categories of protein secondary structure based on the sequence features and the profile features of the protein.

In this paper, we have proposed a new deep convolutional neural network namely PS8-Net which can improve the accuracy of eight-state protein secondary structure prediction. Our proposed architectures verified on several public datasets such as CullPdb6133, CullPdb6133-filtered, CB513, CASP10, and CASP11. Experiment outcomes manifest that our proposed PS8-Net outperforms all the state-of-art methods on CullPdb-6133, CB513, CASP10, and CASP11 to the best of our knowledge.  

The rest of the paper is organized as follows: in section 2, we discuss several related works and some state-of-the-art architectures. The datasets and the methodology are demonstrated in section 3. Section 4, illustrates the performance analysis of the experiment. Finally, we conclude the paper in section 5.

\section{RELATED WORK}
In 1951, hydrogen-bonding patterns gets predicted by Pauling and Corey\cite{pauling1951structure,pauling1951configurations} from $\alpha$-helices and $\beta$-sheets. This finding opened the door to the protein secondary structure research. Although in 1958, the first protein secondary structure (PSS) prediction discovered by X-ray crystallography\cite{kendrew1958three}. The protein secondary structure prediction approach splits into three different generations\cite{rost2001protein}. Secondary structure prediction accomplishes from amino acid residues' statistical propensities of the protein sequence in the first generation\cite{scheraga1960structural,finkelstein1971statistical,chou1974prediction}, and the most popular method of this was Chou–Fasman method\cite{chou1974prediction}. The second-generation techniques employed an sliding window of neighboring residues (LIM method\cite{lim1974algorithms}, GOR method\cite{Garnier1978AnalysisOT}) and several theoretical algorithms like nearest neighbor algorithm\cite{Yi1993ProteinSS}, graph theory\cite{Mitchell1990UseOT}, neural network-based techniques\cite{Holley1989ProteinSS}, statistical information\cite{Garnier1978AnalysisOT,Kabat1973TheIO}, and logic-based machine learning approach\cite{Muggleton1992ProteinSS}. The third-generation methods mostly based on evolutionary information obtained from the alignment of multiple homologous sequences\cite{Rost1993ImprovedPO}. The most notable enhancement in protein secondary structure prediction achieved by Position-specific scoring matrix (PSSM) based on PSI-BLAST\cite{altschul1997gapped} reflects evolutionary information. Many novel computational algorithms have been proposed during this time. For instance, Bayesian or hidden Markov model\cite{Yao2007ADB}, conditional random fields for combined prediction\cite{Liu2004ComparisonOP}, and support vector machines\cite{Ward2003SecondarySP}. Among all of these latest models, deep learning-based approaches provide superior reported accuracy\cite{zhou2018cnnh_pss,guo2018protein,guo2019deepaclstm,wang2016protein,Ma2018ProteinSS}. Specifically, in this period, the Q3 accuracy of protein secondary structure reached above 80$\%$. However, the traditional methods are not able to produce a satisfactory Q8 accuracy rate, since the classification of the eight-state protein secondary structure is more complicated and challenging\cite{zhou2014deep,fang2018mufold,Snderby2014ProteinSS}.

In recent years, deep neural network (DNN) models have become established tools for the representation learning of many distinct data\cite{Angermueller2016DeepLF,Gawehn2016DeepLI,Asgari2015ContinuousDR}. By incorporating evaluation information, deep neural networks (DNNs) also attained significant increase in accuracy on the eight-state (Q8) protein secondary structure prediction\cite{fang2018mufold}. Wang et al.\cite{wang2016protein} integrating shallow neural network and Conditional Random Fields (CRF) and proposed Deep Convolutional Neural Fields (DeepCNF) architecture. Their experiment result shows that DeepCNF can achieve almost 84$\%$ Q3 accuracy and 72$\%$ Q8 accuracy on the CASP, and CAMEO test proteins, respectively. Based on a semi-random subspace method (PSRSM) and data partition in\cite{Ma2018ProteinSS}, Ma et al. have presented a novel architecture that utilizes Support Vector Machines (SVMs) as the base classifier. They achieved 86.38$\%$, 84.53$\%$, 85.51$\%$, 85.89$\%$, 85.55$\%$, and 85.09$\%$ Q3 accuracy respectively on 25PDB, CB513, CASP10, CASP11, CASP12, and T100 datasets. A new deep learning architecture suggested by Zhou et al.\cite{zhou2018cnnh_pss}, namely CNNH-PSS, where they were using multi-scale CNN with the highway to predict the accuracy of PSS. They accomplished 74.0$\%$, and 70.3$\%$ Q8 accuracy on CullPdb6133, and CB513, respectively. Guo et al.\cite{guo2019deepaclstm} proposed the DeepACLSTM model to predict eight-class secondary structure from profile features and sequence features. In DeepACLSTM, they combined ACNNs (asymmetric convolutional neural networks) with BLSTM (bidirectional long short-term memory) neural networks together and found 74.2$\%$, 70.5$\%$, 75.0$\%$, 73.0$\%$ Q8-accuracy respectively on CullPdb6133, CB513, CASP10, and CASP11. Fang et al.\cite{fang2018mufold} provide a Deep inception-inside-inception (Deep3I) model to predict the Q3 and Q8 accuracy for PSS, and later they implemented this as a software tool called MUFOLD-SS. MUFOLD-SS obtained 70.63$\%$, 76.47$\%$, 74.51$\%$, 72.1$\%$ Q8 accuracy respectively on CB513, CASP10, CASP11, and CASP12. Zhang et al. presented a novel method based on a convolutional neural network (CNN), bidirectional recurrent neural network, and residual network to enhance the prediction accuracy for the eight-state protein secondary structure (Q8)\cite{Zhang2018PredictionO8}, and attained 71.4$\%$ accuracy on the benchmark CB513 dataset. In\cite{Busia2016ProteinSS}, Busia et al. suggested an ensemble method consisting of various popular deep neural architectures with BatchNomalization to predict the Q8 accuracy of PSS and reaches 70.6$\%$ eight-state accuracy on CB513 dataset.

\section{Methodology and Datasets}
\begin{figure*}[htbp]
\centerline{\includegraphics[width=18cm]{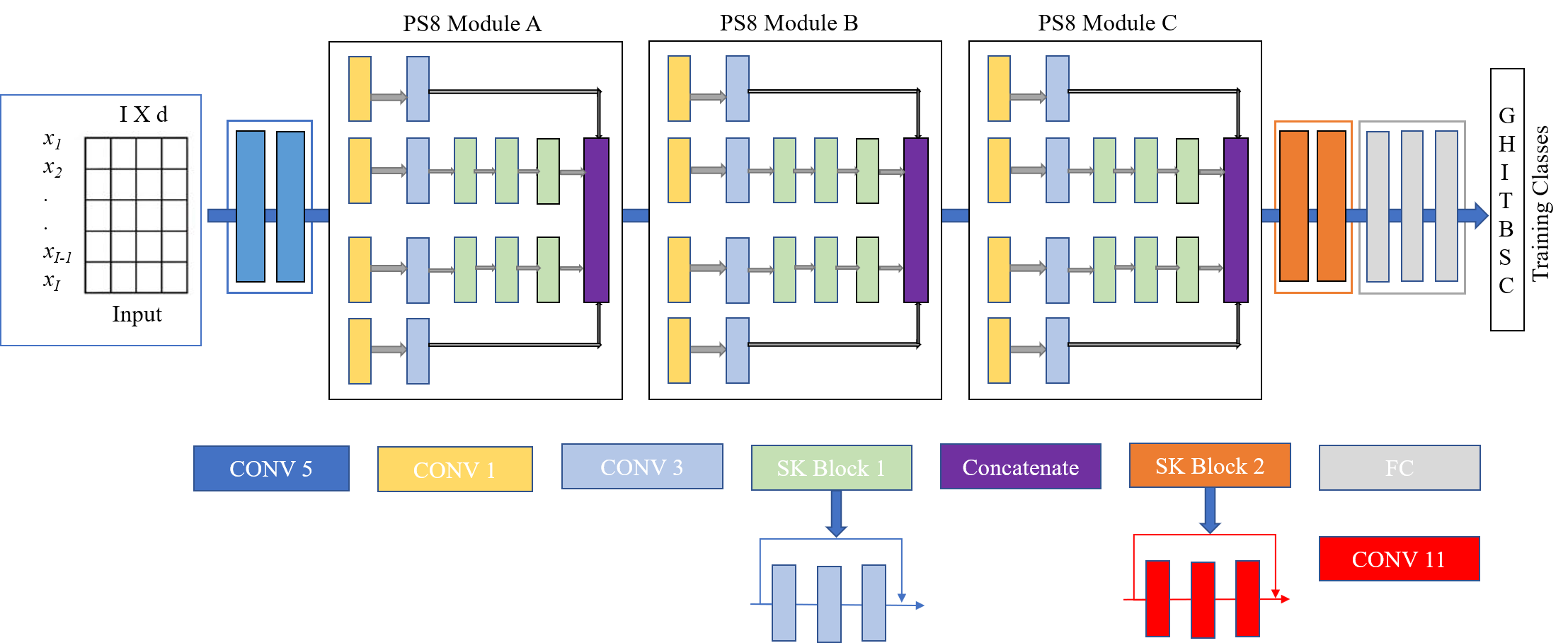}}
\caption{The schematic of our proposed PS8-Net model. It is comprised of three PS8 modules with two skip connection blocks. The network has various types of convolution operation layers and three fully connected layers.}
\label{fig2}
\end{figure*}

In this section, we provide a detailed explanation about the feature representation of the model, our proposed algorithm, and the public datasets used for this study. In our proposed PS8-Net, we have several different convolutional layers with different convolution operations, three PS8 modules, skip connections, and fully connected layers. PS8 module and skip connection attend in extracting the long-term inter-dependencies by higher layers, acquiring local contexts from the lower layer, and achieving global information during classification tasks.  

\subsection{Input Feature Representation of PS8-Net}
For a deep neural network, initial feature representation is essential to predict secondary structure. In our method, we cautiously design feature representation, which consists of sequence features and rich information of profile features. A protein with $ I$ amino acid residues can denote as $X=x_{1},x_{2},x_{3},..,x_{I-1},x_{I}$, where $x_n\epsilon\mathbb{R} ^d$ is represented as $d $-dimensional feature vector of the $n^{\text{th}}$ residue. The eight states of secondary structure label of this specific protein (X) can be formulated as, $S=s_{1},s_{2},s_{3},..,s_{I-1},s_{I}$. Both sequence features and profile features are needed to encode the feature vector $x_n$. In this study, 21-dimensional sequence features have been used to encode the types of target residues. 21-dimensional profile features acquired from the PSI-BLAST log file\cite{altschul1997gapped} and later logistic function\cite{jones1999protein} used to re-scale it. Sequence features are represented as a 21-dimensional one-hot vector, in which all elements are "0" except only one "1". Furthermore, the profile features have dense representation in this study. Therefore, to maintain the consistency in all feature representations of proteins, we apply an embedding operation to convert sparse sequence features into the dense vector. This embedding technique maps the 21-dimensional dense vector from the 21-dimensional sparse vector. Finally, a 42-dimensional initial protein feature can be obtained by concatenating both the dense representation of the sequence feature and the profile feature.

The input of the architecture is the $I $X$ d$ matrix where the length of the input sequence defines as $I$, and $d$ is the dimension of vectors. For input, feature vector of the $n^{\text{th}}$ residue in protein can represent as $x_n\epsilon \mathbb{R} ^d$. Here, the value of $I=700$ and $d=42$, and the output of this network is secondary structure labels. Here, $I$ can be encoded as $I $X$ d$ matrix $ \ {x}_{1:I}= [x_{1}, x_{2},..., x_{I}]^T$, where $I$ is the length of protein and d denote the number of input features utilized to encode residues.

\subsection{CNN with PS8 Module}
In a deep convolutional neural network (CNN), a kernel or sliding window can scrutinize the local patch from the provided input sequences. Additionally, in the local patch, this kernel operation extract interdependence among the amino acid residues. Thus, to get local dependencies of adjoining amino acids throughout the system, including our proposed PS8 modules, we apply four different kernel sizes: 1, 3, 5, 11. In Figure 1, "CONV 1", "CONV 3", "CONV 5", and "CONV 11" denote convolutional layers with kernel size 1, 3, 5, and 11, respectively. Moreover, we apply the Rectified Linear Unit (ReLU) with all the convolutional layers to extract long-term inter-dependencies between residues with more long-distance.
 
In the output convolutional layer, to retain the same height as the input, to the head and the tail of the input $x_{1 : I}$ respectively, we require to pad $[L/2]$ and $[(L-1)/2]$. Here, $L$ denote the length of kernels in the specific convolutional layers. Our PS8 modules and two convolutional layers of the "CONV5" operation hold $n$ convolutional layers. In the $(l-1)$ convolutional layer, the convolutional operation $ {w}_{k}^{(l-1)}\epsilon\mathbb{R}^{LXd}$ of $k^{\text{th}}$ employed on protein fragment $x_{i:i+L-1}$ is denoted as:
 
 \begin{equation}
y_{k,j}^{l-1} = ReLU( w_{k}^{l-1} . x_{j:j+L-1} + b_{k}^{l-1})
\end{equation}

 Here L is the kernel extent, $b_{k}^{l-1}$ is the bias term of a particular kernel $k^{\text{th}}$, $ReLU$ is the activation function which we have utilized in all of our PS8 module, the protein fragment $x_{j:j+L-1}$ can be formulated as $x_{j},x_{j+1},x_{j+2},...,x_{j+L-1}$. Moreover, after applying convolution operation of the $k^{\text{th}}$ kernel on all input sequences incorporate length $L$ of the padded input, a feature vector is acquired:
 
 \begin{equation}
y_{k}^{l-1} = [y_{k,1}^{l-1},y_{k,2}^{l-1},y_{k,3}^{l-1},...,y_{k,I}^{l-1}]^T
\end{equation}

 If we have $p$ numbers of kernels, we can produce $p$ numbers of the feature vector. Next, we can obtain a new feature matrix from a particular convolutional layer with $IXp$ dimension by concatenating $p$ feature vectors. 
 
 \begin{equation}
y^{l-1} = [y_{1}^{l-1},y_{2}^{l-1},y_{3}^{l-1},...,y_{I}^{l-1}]
\end{equation}

This produced feature matrix from a convolutional layer will be used as an input for the next convolutional layer. There are four series of the convolutional layer in the PS8 module. The first and fourth series contains "CONV 3" followed by the "CONV 1" layer. Furthermore, the second and third series includes three extra "SK Block 1" with these "CONV 1" and "CONV 3" layers. These four series follow the same methods to generate the outcome. Initially, for the first and fourth series there are two convolutional layers in a PS8 module, and $A_l$ is the kernel and bias of the $l^{\text{th}}$ convolutional layer, then the output  for these two series will be:
 
 \begin{equation}                                                         
y_1 = f_{A_2}^{2}(f_{A_1}^1(x(1:L)))
\end{equation}
 \begin{equation}                                                         
y_4 = f_{A_2}^{2}(f_{A_1}^1(x(1:L)))
\end{equation}

If we assume that there are $l$ number of convolutional layers in each of the second and third series then the outcome will be:

\begin{equation}                                                         
y_2 = f_{A_{l}}^{l}(f_{A_{l-1}}^{l-1}(...f_{A_1}^1(x(1:L))))
\end{equation}
\begin{equation}                                                         
y_3 = f_{A_{l}}^{l}(f_{A_{l-1}}^{l-1}(...f_{A_1}^1(x(1:L))))
\end{equation}
 
Finally, we can concatenate the output of these four series of convolutional layers to get the final output of the PS8 module. 

\begin{equation}                                                         
Y = CONCAT[y_1, y_2, y_3, y_4]
\end{equation}
Here, Y stands for the final output from the PS8 module. The number of hidden units throughout the PS8 Module A is 256. Both the PS8 Module B and PS8 Module C contain 128 hidden units in the CNN layers. Finally, throughout these three modules, we apply a 0.25 dropout rate to prevent our neural network from overfitting. 
 
\begin{figure}[htbp]
\centerline{\includegraphics[width=8cm]{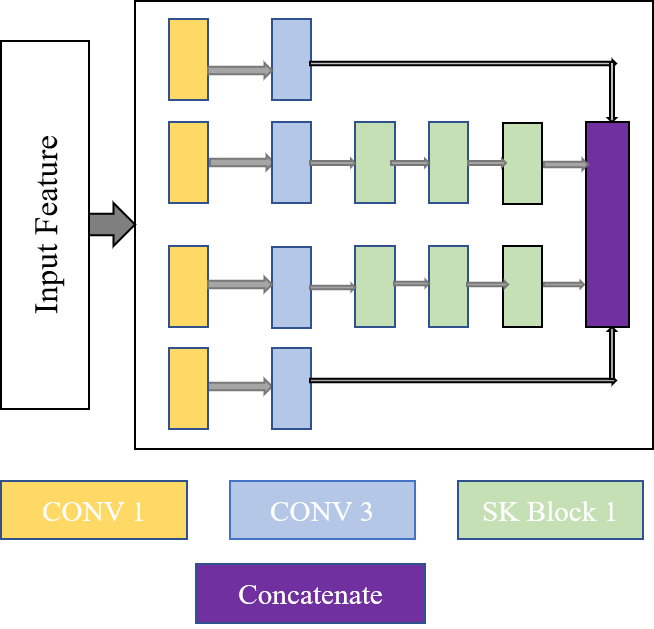}}
\caption{PS8 module with four main operations. The golden rectangular bar "CONV 1": convolutional layers with kernel size 1, the light blue rectangular bar "CONV 3": convolutional layers with kernel size 3, the light green rectangular bar "SK Block 1": skip connection operation, and the purple rectangular bar "Concatenate": accomplishes feature concatenation.
}
\label{fig1}
\end{figure}

\subsection{Skip Connection}

A convolutional neural network (CNN) with a high number of convolutional layers has the capacity to extract long-term interdependencies between amino acid residues with more long sequences. However, when we enhance the number of CNN layers, the network will drop the important local contexts information learned during the early training process. Besides, the exploding/vanishing gradients problem is a colossal issue in the convolutional neural networks (CNN), which is integrated with a large amount of layers \cite{srivastava2015training}. Therefore, to tackle this issue, in our proposed system, we employ "SK Block 1" and "SK Block 2" which are based on the skip connections mechanism\cite{tong2017image} \cite{he2016deep}. Previously, this mechanism was used in several popular image recognition networks such as ResNet \cite{he2016deep}. The PS8 module and "skip connection" help us achieve local contexts from lower layers for protein secondary structure prediction. 

In PS8-Net, we have two types of "skip connection" blocks: "SK Block 1" and "SK Block 2". Both of these blocks have the same skip connection mechanism except the kernel size. In "SK Block 1" the kernel size is 3, whereas in the "SK Block 2" the kernel size is 11. Additionally, these skip connection blocks have three convolutional layers with BatchNormalization and $ReLU$ activation function. The SK blocks have the similar input activation dimensions ($a^{[l]}$) and the output activation dimension ($a^{[l+3]}$). In these blocks, “skip connection” has been utilized to backpropagate the gradient to the initial layers of these blocks. Besides, with this “skip connection” any SK Blocks can suitably learn the identity mapping\cite{he2016deep}. The output of the SK Blocks $y^{l+3}$ is based on two issues: the knowledge gain from $l^{\text{th}}$ layer and the outcome of convolutional kernels of the present layer. 
\begin{equation}                                                         
y^{l+3}=ReLU(f_{A_{l+3}}^{l+3}(y^{l+2}) + y^l)
\end{equation}
Here, $f_{A_{l+3}}^{l+3}$ denoted as the convolution operation of the last layer of SK Blocks. Furthermore, for this convolution operation, the function takes the feature map from the previous layer. Finally, to get the final output, input feature map of this block or, $y^l$ (value from the skip connection) has been added. 
\subsection{PS8-Net Architecture}
We used the ReLU activation function in all convolutional layers. In the PS8-Net, the first two convolutional layers, namely "CONV 5" have 512 hidden units and "SK Block 2", has 128 hidden units. In the classifier part, we implement three fully connected ("FC") layers with 512, 256, and 8 (because we have eight classes) hidden units, respectively. The first two FC layers contain the ReLU activation function with dropout rate 0.50, and the last one has the SoftMax activation function to predict eight states of protein secondary structure. 
\begin{equation}                                                         
F_j=SoftMax(w.F_{l}^{j-1}+b)
\end{equation}
Where $F_j$ is its predicted protein secondary structure from the last fully connected ("FC") layer. Moreover, $w$, and $b$ indicate the weight and bias, respectively. Lastly, $F_{l}^{j-1}$ is denoted as feature vector of the $l^{\text{th}}$ outcome by the  previous FC layer ($j-1^{\text{th}}$). 

\subsection{Datasets}
Here, we utilize five different datasets, namely, CullPdb6133, CullPdb6133-filtered, Cb513, Casp10, and Casp11. Among these five datasets, CullPdb6133, and CullPdb6133-filtered are used for training. Furthermore, CB5133, Casp10, Casp11, and 272 protein sequences of CullPdb6133 are employed for the testing.

1) CullPdb6133: CullPdb6133\cite{Wang2003PISCESAP} dataset is a non-homologous protein dataset provided by PISCES CullPDB with the familiar secondary structure for protein. This dataset contains a total of 6133 protein sequences, in which 5600 ([0:5600]) protein samples are considered as the training set, 272 protein samples [5605:5877] for testing, and 256 proteins samples ([5877,6133]) regarded as the validation set. Moreover, CullPdb6133 (non-filtered) dataset has 57 features, such as amino acid residues (features [0:22]), N- and C- terminals (features [31,33]), relative and absolute solvent accessibility ([33,35]), and features of sequence profiles (features [35:57]). There are secondary structure notations (features [22:31]) that have been used for labeling. This CullPdb dataset is publicly obtainable from\cite{zhou2014deep}.

2) CullPdb6133-filtered: There exists redundancy between the CB513 dataset and the CullPdb6133 dataset. Therefore, in the CullPdb6133-filtered version, 5534 protein sequences were created by detaching those sequences with over 25$\%$ resemblance between CB513 and CullPdb6133. In this dataset, we take a 5234 protein sequence randomly picked for training and the remaining 300 protein sequences for validation.

3) CB513: CB513 dataset contains 514 protein sequences, and this dataset widely utilized to compare the performance of protein secondary structure prediction. Here, we use this dataset for evaluating our testing result after training with CullPdb6133-filtered. This dataset can be accessed from\cite{zhou2014deep,Cuff1999EvaluationAI}.

4) Casp10 and Casp11: Casp10 and Casp11\cite{zhou2018cnnh_pss,li2016protein} respectively contain 123 and 105 protein sequences. Since 1994, the Critical Assessment of protein Structure Prediction or CASP used for protein structure prediction worldwide. The bioinformatics community has hugely exploited these datasets. After training our models with CullPdb6133-filtered, we were using the CASP dataset for testing.

In order to more properly encoding and processing of protein sequences, the length of all protein sequences are normalized to 700, according to \cite{zhou2018cnnh_pss}. In CullPdb6133 and CullPdb6133-filtered datasets, we have reshaped them from 6133 proteins x 39900 features and 5534 proteins x 39900 features into 6133 proteins x 700 amino acids x 57 features, and 5534 proteins x 700 amino acids x 57 features respectively. where, 700 indicates the peptide chain, and 57 signifies the features for each amino acid. We use sequence features and profile features (total of 42 features) for training input. Besides, the output is a protein secondary structure labelling.

\section{Experiment Results}
Here, we focus on the overall accuracy of eight classes of protein secondary structure prediction. This section presented broad experiment outcomes for Q8 protein secondary structure prediction of our proposed PS8 network on the aforementioned public datasets and compared the performance with various popular state-of-the-art methods. At first, we delineate the implementation strategy that is followed during the training process of our network. Then, we analyze the architecture with the CB513 dataset to justify the input feature selection. After that, several different combinations are tried for this proposed architecture to decide what would be the suitable strategies in terms of getting superior test accuracy. Finally, we present the comparison result with existing popular systems.

\subsection{Implementation Details}
We trained our proposed PS8-Net for 120 epochs with Adam optimization algorithm. Furthermore, in the whole training, the learning rate is $2e-4$, the mini-batch size is 64, and we apply cross-entropy loss as the loss function. We have used one callback function to reduce the learning rate factor by $\sqrt{0.1}$ if learning stagnates for seven epochs, and the lower bound for the learning rate is $0.5e-5$. 
\begin{equation}
l_{new} = l * \sqrt{0.1}
\end{equation}

Where, $l$= present learning rate, and $l_{new}$= new learning rate.

\subsection{Determine the Input Features}

\begin{figure}[htbp]
\centerline{\includegraphics[width=9cm]{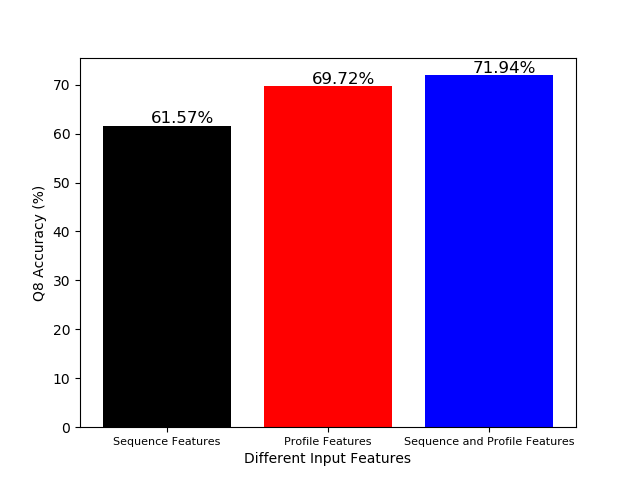}}
\caption{Performance analysis for PS8-Net on various input features
}
\label{fig1}
\end{figure}
We inspect whether both the profile features and sequence features are essential for our experiment to predict the eight-state PSS or not. We utilize the CullPDB6133-filtered dataset for training and CB513 for testing to run three operations. In the first experiment, we evaluate our proposed PS8-Net with only sequence features and achieve 61.57\% Q8 accuracy. Next, in the second experiment, we attain 69.72\% Q8 accuracy when we only use profile features. Finally, we utilize both sequence and profile features to evaluate the proposed architecture and gain 71.94\% Q8 accuracy. From Figure 3, we can exhibit that our proposed network acquire superior results when we take both sequence features and profile features as the input.  

\begin{table}[!tbp]
\caption{Eight-state accuracy for PS8-Net with different number of PS8 modules and best outcomes marked in \textbf{bold}}
\begin{center}
\begin{tabular}{|p{1.3cm}|p{1.4cm}|p{1.3cm}|p{1.3cm}|p{1.3cm}|}
\hline
PS8 Module Numbers & Test Accuracy on CullPdb6133 (\%) & Test Accuracy on CB513 (\%) & Test Accuracy on CASP10 (\%) & Test Accuracy on Casp11 (\%)\\

\hline
1 & 72.97 & 68.39 & 73.51 & 70.24 \\

\hline
2 & 75.88 & 70.61 & 75.70 & 72.93 \\

\hline
\textbf{3} & \textbf{76.89} & \textbf{71.94} & \textbf{76.86} & \textbf{75.26} \\

\hline
4 & 75.19 & 70.04 & 74.85 & 73.63 \\

\hline
5 & 74.34 & 69.47 & 73.79 & 72.54 \\
\hline
\end{tabular}
\label{tab1}
\end{center}
\end{table}

\subsection{Ablation Study}
We utilize several datasets to evaluate the performance of the proposed architecture. Firstly, train on 6128 protein sequences of CullPdb6133, validated on 256 protein sequences and tested on the remaining 272 protein sequences. Next, picked 5234 protein sequence randomly from the CullPdb6133-filtered dataset for training, validated on the remaining 300 sequences, and tested on Cb513, Casp10, Casp11.

In TABLE \Romannum{1}, we have displayed the outcome for the several different combinations of our proposed PS8-Net. We inspect the accuracy by increasing or reducing the PS8 module numbers in the model. We achieve superior test accuracy for different datasets when there are three PS8 modules in PS8-Net. Besides, we employ different sets of hidden units for the PS8 modules, though the network manifests superior test accuracy when the hidden units are 256, 128, 128 for module A, module B, and module C, respectively. Moreover, with this combination, a dropout rate of 0.25 in the modules gives us the best accuracy. During the training period, we use many different combinations; however, in TABLE \Romannum{1}, we displayed only the best five outcomes. 

Furthermore, in TABLE \Romannum{2}, we show that skip connection has a significant impact on achieving the high accuracy. We obtain the best result by adding both the "SK Block 1" and "SK Block 2" in our network.

\begin{table}[!tbp]
\caption{Influence of skip connection on PS8-Net and the best results highlighted in \textbf{bold}}
\begin{center}
\begin{tabular}{|p{1.7cm}|p{1.4cm}|p{1.1cm}|p{1.3cm}|p{1.1cm}|}
\hline
Setups & Test Accuracy on CullPdb6133 (\%) & Test Accuracy on CB513 (\%) & Test Accuracy on CASP10 (\%) & Test Accuracy on Casp11 (\%)\\

\hline
PS8-Net (without any skip Connection) & 69.52 & 63.49 & 70.01 & 68.73 \\

\hline
PS8-Net + with only SK Block 1 & 73.24 & 69.19 & 74.63 & 73.14 \\

\hline
\textbf{PS8-Net + SK Block 1 + SK Block 2} & \textbf{76.89} & \textbf{71.94} & \textbf{76.86} & \textbf{75.26} \\

\hline

\end{tabular}
\label{tab1}
\end{center}
\end{table}

\subsection{Comparison with the State-of-the-Art Architectures}
Protein secondary structure prediction is a vital issue to solve in bioinformatics because of drug discovery\cite{adamczak2004accurate,baldi1999exploiting}, and analyzing protein function. There are several popular methods proposed in the past to predict eight-state protein secondary structure. Recently, most of the state-of-the-art models are based on deep learning. In this part, we exhibit the comparison result of our proposed model and the following state of the art architectures.

1) SSPro-8: In \cite{Pollastri2001ImprovingTP}, Pollastri et al. have utilized PSI-BLAST-derived profiles and ensemble the Bidirectional RNN model to enhance the eight-state prediction accuracy of protein secondary structure.

2) MUFOLD-SS: A new deep neural network proposed to map the protein secondary structure where Fang et al. \cite{fang2018mufold} introduce a nested inception network. Like us, they also used a deep convolutional neural network (DCNN) approach to predict the eight-state PSS. 

3) CNNH-PSS: Zhou et al. \cite{zhou2018cnnh_pss} have proposed a deep learning based approach, namely CNNH-PSS, by employing a multi-scale convolutional network with the highway technique to predict Q8 PSS.

4) DeepACLSTM: In \cite{guo2019deepaclstm}, we find a model where Guo et al. provide an asymmetric CNN (ACNN) merge with Bidirectional LSTM to predict protein secondary structure. They exploit sequence features and profile features to find 8 class of PSS.

5) 2C-BRNN: To improve the accuracy of eight-state PSS Guo et al. \cite{guo2018protein} have proposed hybrid deep learning architectures that are based on 2D Convolutional Bidirectional RNN (2C-BRNN). They have used Bidirectional Long Short term Memory and Bidirectional Gated Recurrent Units to construct four different models. Their proposed architectures are: 2DCNN-BGRUs,2DConv-BGRUs,2DCNN-BLSTM, and 2DConv-BLSTM.

6) DeepCNF: In \cite{wang2016protein}, Wang et al. extend their CNF model using deep learning techniques to make it more powerful. It is an integration between shallow neural networks and Conditional Random Fields (CRF).

7) GSN: To predict protein secondary structure, Zhou et al. \cite{zhou2014deep} have presented a supervised Generative Stochastic Network (GSN) with deep hierarchical representations.

8) DCRNN: Li et al. \cite{li2016protein} have proposed an end-to-end deep neural network that predicts eight-class PSS from integrated local and global contextual features.


\begin{table}[htbp]
\caption{Comparison of the Q8(\%) accuracy of PS8-Net and some existing benchmark methods on CullPdb6133 dataset}
\begin{center}
\begin{tabular}{|c|c|}
\hline
Method  & Q8(\%) Accuracy After Tested on CullPdb6133 \\

\hline
SSPro-8 \cite{Pollastri2001ImprovingTP} & 66.6\\

\hline
GSN \cite{zhou2014deep}&72.1\\

\hline
DeepCNF \cite{wang2016protein}&75.2 \\

\hline
CNNH-PSS \cite{zhou2018cnnh_pss}&74.0\\

\hline
DCRNN \cite{li2016protein}&73.2 \\

\hline
2DConv-BGRUs \cite{guo2018protein} & 74.9\\
\hline
2DConv-BLSTM \cite{guo2018protein}& 75.7\\

\hline
2DCNN-BGRUs \cite{guo2018protein}& 73.5\\

\hline
2DCNN-BLSTM \cite{guo2018protein}& 74.3\\

\hline
 \textbf{PS8-Net (ours)}& \textbf{76.9}\\
\hline

\end{tabular}
\label{tab1}
\end{center}
\end{table}

In TABLE \Romannum{3}, we have displayed the comparison result of the Q8 accuracy of the protein secondary structure of our proposed PS8-Net on CullPdb6133 and several popular benchmark models, which are trained on 6128 protein sequences of CullPdb6133 and tested on the 272 protein sequences from the same dataset. According to TABLE \Romannum{3}, our proposed PS8-Net attains the highest Q8 accuracy that exceeds the accuracy of all the benchmark architectures for the CullPDB6133 dataset. PS8-Net exhibits 76.9$\%$ overall eight-state accuracy, which is 1.2$\%$ and 1.7$\%$ more than 2DConv-BLSTM\cite{guo2018protein} and DeepCNF model, respectively.

Next, we have trained our model on the CULLPdb6133-filtered dataset and tested on CB513, CASP10, and CASP11 datasets. Then, we compare several benchmark models with our PS8-Net by eight-state (Q8) accuracy of PSS. According to TABLE \Romannum{4}, our proposed architecture PS8-Net presents 71.94$\%$, 76.86$\%$, 75.26$\%$ Q8 accuracy for CB513, CASP10, and CASP11 datasets, respectively. Moreover, PS8-Net shows superior overall accuracy than any other state-of-the-art models for protein secondary structure on CB513, CASP10, CASP11, to the best of our knowledge.

\begin{table}[!tbp]
\caption{Comparison of Q8 ($\%$) accuracy for PSS of PS8-Net and some state-of-the-art method on the CB513, CASP10, and CASP11 dataset and best results marked in \textbf{bold}}
\begin{center}
\begin{tabular}{|p{2.6cm}|p{1.5cm}|p{1.5cm}|p{1.5cm}|}
\hline
Method & Accuracy on CB513 (\%) & Accuracy on CASP10 (\%) & Accuracy on Casp11 (\%)\\

\hline
SSPro-8 \cite{Pollastri2001ImprovingTP}  & 63.5 & 64.9 & 65.6 \\

\hline
DeepCNF \cite{wang2016protein} & 68.3 & 71.8 & 72.3 \\
\hline
DCRNN \cite{li2016protein} & 69.4 & - & - \\
\hline
CNNH-PSS \cite{zhou2018cnnh_pss} & 70.3 & - & -  \\

\hline
DeepACLSTM \cite{guo2019deepaclstm} & 70.5 & 75 & 73 \\
\hline
MUFOLD-SS \cite{fang2018mufold} & 70.63 & 76.47 & 74.51 \\
\hline
2DConv-BGRUs \cite{guo2018protein} & 69.0 & 72.2 & 71.0 \\
\hline
2DConv-BLSTM \cite{guo2018protein} & 70.2 & 74.5 & 72.5 \\
\hline
2DCNN-BGRUs \cite{guo2018protein} & 68.7 & 72.1 & 71.7 \\
\hline
2DCNN-BLSTM \cite{guo2018protein}& 70.0 & 74.5& 72.6 \\

\hline
 \textbf{PS8-Net (ours)}& \textbf{71.94} & \textbf{76.86} & \textbf{75.26}\\
\hline

\end{tabular}
\label{tab1}
\end{center}
\end{table}

\section{Conclusion}
In this paper, we have introduced a new deep convolutional neural architecture to predict the eight-state protein secondary structure. Our proposed network (PS8-Net) is trained and tested on five most popular publicly available benchmark datasets. CullPdb6133, and CullPdb6133-filtered are utilized in training. Then, the network is tested on CullPdb6133, CB513, CASP10, and CASP11 datasets. We have proposed a new PS8 module in the network, which is applied with skip connection for the long-term inter-dependencies extraction from higher layers, obtaining local contexts in earlier layers and achieving global information during secondary structure prediction as well. Our proposed PS8-Net attains 76.89$\%$, 71.94$\%$, 76.86$\%$, and 75.26$\%$ Q8 accuracy respectively on benchmark datasets, which enables the efficient processing of both local and global interdependencies between amino acids to make a precise prediction of each category. PS8-Net experiment results exhibit that it surpasses other state-of-the-art methods on the aforementioned benchmark datasets, to the best of our knowledge.

\bibliographystyle{IEEEtran}
\bibliography{ref}

\end{document}